\documentclass[10pt,twocolumn,letterpaper]{article}

\usepackage[pagenumbers]{cvpr} % To force page numbers, e.g. for an arXiv version
\usepackage{lineno}
\usepackage{graphicx}
\usepackage{amsmath}
\usepackage{amssymb}
\usepackage{booktabs}
\usepackage{multirow}
\usepackage{booktabs}
\usepackage{algorithm}
\usepackage{algorithmicx}
\usepackage{mathtools}
\usepackage{adjustbox}
\usepackage{import}
\usepackage[noend]{algpseudocode}
\usepackage{caption}

\def\real{\mathbb{R}}
\def\loss{\mathcal{L}}

\usepackage{multirow}
\usepackage{rotating}
\usepackage{xcolor}
\usepackage[dvipsnames]{xcolor}
\usepackage{colortbl}

\definecolor{dark_green}{rgb}{0, 0.5, 0}
\definecolor{dark_red}{rgb}{0.8, 0.2, 0.2}
\definecolor{soft_red}{rgb}{1.0, 0.4, 0.4}

\definecolor{cvprblue}{rgb}{0.21,0.49,0.74}
\usepackage[pagebackref,breaklinks,colorlinks,allcolors=cvprblue]{hyperref}

\title{Reconstructing Humans with a Biomechanically Accurate Skeleton}

\author{
    Yan Xia$^{1,2}$
    \quad
    Xiaowei Zhou$^{2}$
    \quad
    Etienne Vouga$^{1}$
    \quad
    Qixing Huang$^{1}$
    \quad
    Georgios Pavlakos$^{1}$
    \\[1.5mm]
    $^{1}$The University of Texas at Austin
    \quad
    $^{2}$Zhejiang University
}

\begin{document}

\twocolumn[{%
\renewcommand\twocolumn[1][]{#1}%
\maketitle
\begin{center}
    \newcommand{\teaserwidth}{\textwidth}
    \vspace{-1.5em}
    \centerline{
        \includegraphics[width=1.0\teaserwidth,clip]{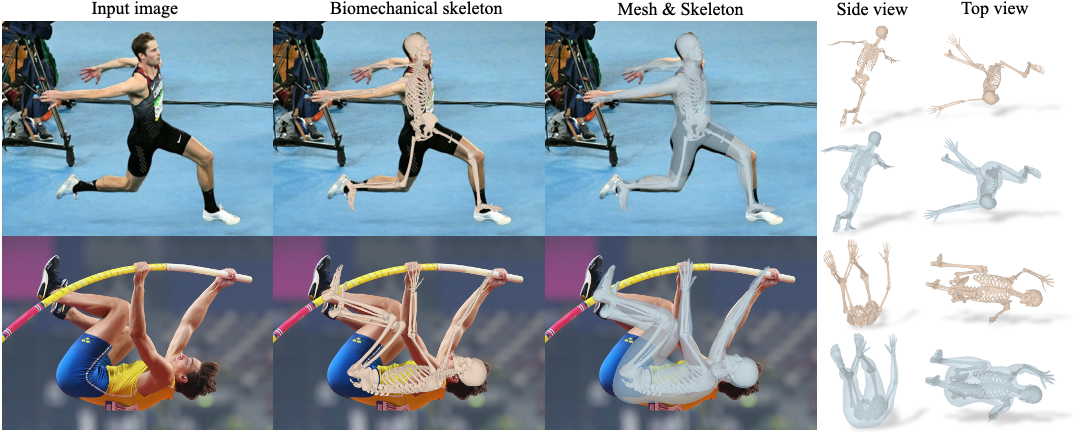}
    }
    \vspace{-0.07in}
    \captionof{figure}{{\bf Human Skeleton and Mesh Recovery (HSMR)}.
    We propose an approach that recovers the biomechanical skeleton and the surface mesh of a human from a single image.
    We adopt a recent biomechanical model, SKEL~\cite{keller2023skin} and train a transformer to estimate the parameters of the model.
    We encourage the reader to see the skeleton and surface reconstructions in our \href{https://isshikihugh.github.io/HSMR/}{project page}.
    }
    \vspace{-0.15in}
\label{fig:teaser}
\end{center}%
}]

%\maketitle
\begin{abstract}

In this paper, we introduce a method for reconstructing 3D humans from a single image using a biomechanically accurate skeleton model.
To achieve this, we train a transformer that takes an image as input and estimates the parameters of the model.
Due to the lack of training data for this task, we build a pipeline to produce pseudo ground truth model parameters for single images and implement a training procedure that iteratively refines these pseudo labels.
Compared to state-of-the-art methods for 3D human mesh recovery, our model achieves competitive performance on standard benchmarks, while it significantly outperforms them in settings with extreme 3D poses and viewpoints.
Additionally, we show that previous reconstruction methods frequently violate joint angle limits, leading to unnatural rotations.
In contrast, our approach leverages the biomechanically plausible degrees of freedom making more realistic joint rotation estimates.
We validate our approach across multiple human pose estimation benchmarks.
We make the code, models and data available at: \url{https://isshikihugh.github.io/HSMR/}

\end{abstract}
\section{Introduction}
\label{sec:intro}

In recent years, there has been remarkable progress in 3D human pose estimation,
with proposed methods reaching the potential that computer vision researchers envisioned by finding applications in diverse fields
such as robotics~\cite{fu2024humanplus,peng2018sfv,li2024okami, radosavovic2024humanoid}, graphics and animation~\cite{weng2022humannerf, zhu2024champ}, and AR/VR~\cite{tome2019xr}.
However, there are fields where these techniques would seemingly be a perfect fit, yet the adoption has been notably limited.
Biomechanics is one such example.
For biomechanics, the requirements are stricter: we need methods that estimate parameters compatible with biomechanical skeletons, respect joint limits, ensure physically plausible motion and return high accuracy estimates.
Unfortunately, most state-of-the-art methods do not satisfy these constraints, and extensive post-processing is required~\cite{pagnon2022pose2sim, uhlrich2023opencap}.
Our goal is to move towards bridging this gap by proposing an approach that can generate predictions aligned with a biomechanically accurate skeleton model.

Currently, the progress in the field of 3D human pose estimation has largely been driven by the use of parametric body models, like SMPL~\cite{loper2015smpl}, SMPL-X~\cite{pavlakos2019expressive} and GHUM~\cite{xu2020ghum}.
These models provide a compact parameterization,
%which has significantly benefited deep learning approaches~\cite{kanazawa2018end, kolotouros2019learning, kocabas2021pare, goel2023humans} by
enabling the direct regression of model parameters from an input image~\cite{kanazawa2018end, kolotouros2019learning, kocabas2021pare, goel2023humans}.
While these parametric models offer plausible surface representations, their
skeleton design is not anatomically accurate.
For instance, the kinematic tree does not align with the actual skeletal structure of the human body~\cite{keller2023skin}.
Moreover, the joints are represented as ball (socket) joints, introducing additional degrees of freedom.
This modeling choice can lead to the prediction of unnatural joint angles, resulting in outputs that are incompatible with biomechanical applications and simulations~\cite{delp2007opensim}.
Consequently, recent advances in 3D human pose estimation have not yet been fully leveraged by biomechanics, which rely on more anatomically accurate models.

The introduction of the SKEL model~\cite{keller2023skin} marked a significant step forward by integrating a biomechanical skeleton with the SMPL surface mesh.
This combination enables compatibility with biomechanical simulation environments, allowing for more anatomically realistic modeling.
However, to fully harness advancements in computer vision, it is essential to develop methods that can accurately estimate the parameters of this model directly from image inputs.

Towards this goal, we propose \textbf{HSMR} (\textbf{H}uman \textbf{S}keleton and \textbf{M}esh \textbf{R}ecovery), a method for reconstructing humans with a biomechanically accurate skeleton from a single image. % in a biomechanically accurate way.
Our method leverages the recently introduced SKEL model~\cite{keller2023skin} and adopts a transformer-based network~\cite{dosovitskiy2020image, xu2022vitpose} to regress the SKEL parameters from image input.
One key challenge is that there is no dataset of images with corresponding SKEL parameters, that could be used for training.
To address this, we perform an initial optimization~\cite{keller2023skin}
to convert the SMPL (pseudo) ground truth of existing datasets to SKEL pseudo ground truth.
While this is a reasonable starting point, the offline conversion is not perfect and can introduce annotation errors.
To ensure high-quality data, we propose an iterative refinement routine during training, which progressively improves the SKEL pseudo ground truth, enabling us to train a more accurate and reliable model.
This refinement is in the spirit of SPIN~\cite{kolotouros2019learning} -- we optimize the SKEL model to align with the ground truth 2D body keypoints, while using the HSMR estimate as an initialization of the optimization.
The result of this fitting is used as pseudo ground truth for future training iterations.

We carefully benchmark HSMR across multiple datasets.
Despite starting without any ground truth SKEL parameters, we demonstrate that our approach matches the performance of state-of-the-art methods, when evaluated on the traditional metrics for 2D/3D joints accuracy.
More importantly, our model has a clear advantage in cases with extreme poses and viewpoints (\ie, yoga postures from the MOYO dataset~\cite{tripathi20233d}), which often lie outside the distribution of standard training data.
This result indicates that the biomechanical skeleton model can be helpful at regularizing the estimated pose.
Furthermore, we show that previous methods based on SMPL parameter regression frequently yield unnatural joint rotations, due to SMPL’s simplified skeleton modeling, which provides more degrees of freedom than a realistic biomechanical model.

To summarize, our contributions are:
\begin{itemize}
    \item We present HSMR, which is, to the best of our knowledge, the first end-to-end approach that can reconstruct humans in 3D from a single image by estimating the parameters of a biomechanical skeleton model, SKEL~\cite{keller2023skin}.
    \item Starting without any paired dataset of images and SKEL ground truth, we show how to generate data to train our model. Additionally, we incorporate a procedure to iteratively refine the quality of the pseudo ground truth.
    \item We demonstrate that our approach can match the performance of the most closely related state-of-the-art method that regresses SMPL parameters~\cite{goel2023humans}, while achieving clear improvements specifically for more challenging cases with extreme poses and viewpoints.
    \item We highlight the limitations of methods regressing parameters of simpler body models (\ie, SMPL), and show how they tend to predict unnatural rotations for the body joints, leading to biomechanically inaccurate results.
\end{itemize}

\section{Related Work}
\label{sec:related}

\begin{figure*}
    \centering
    \small
    \includegraphics[width=1.0\linewidth]{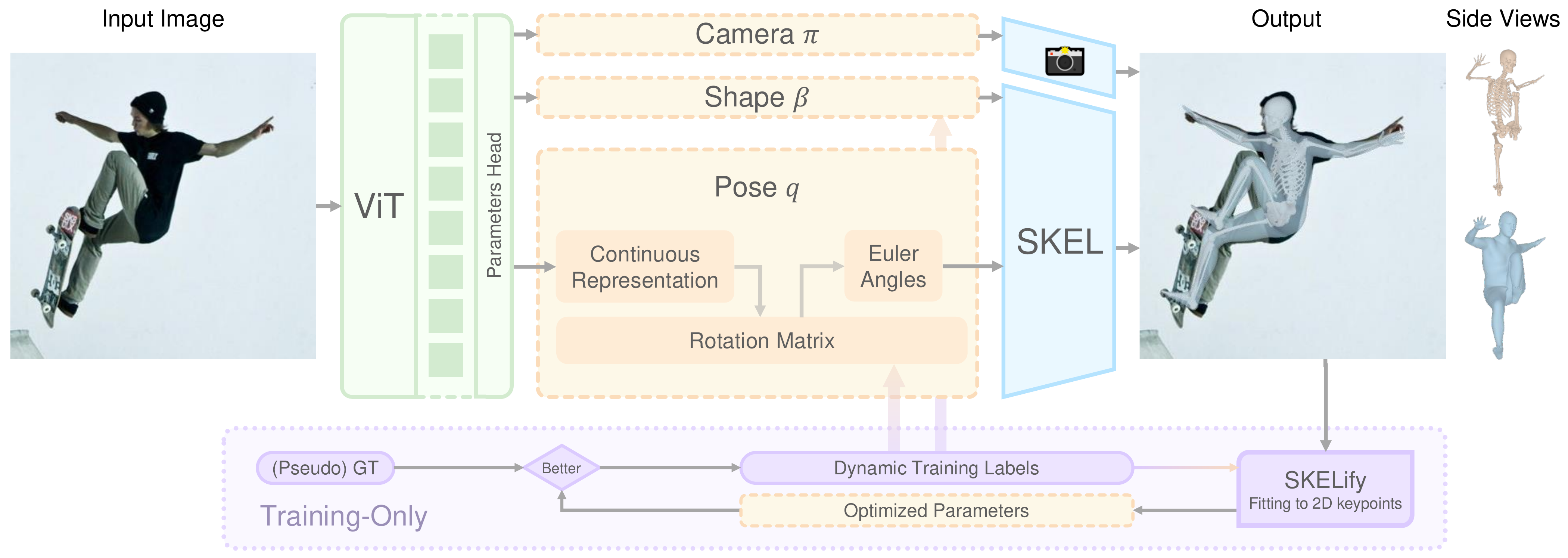}
    \vspace{-1.5em}
    \caption{{\bf Overview of our HSMR approach.}
    A key design choice of HSMR is the adoption of the SKEL parametric body model~\cite{keller2023skin} which uses a biomechanically accurate skeleton.
    We employ a transformer-based architecture that takes as input a single image of a person and estimates the pose $q$ and shape parameters $\beta$ of SKEL, as well as the camera $\pi$.
    During training, we iteratively update the pseudo ground truth we use to supervise our model, aiming to improve its quality.
    For this, we optimize the HSMR estimate to align with the ground-truth 2D keypoints (SKELify).
    The output parameters of the optimization are used in future training iterations as supervision target.
    }
    \label{fig:approach}
    \vspace{-1.5em}
\end{figure*}

\noindent
\textbf{Human Body Models.}
A lot of the recent progress in pose estimation can be attributed to the access to simple, yet realistic models of the human body.
SCAPE~\cite{anguelov2005scape} was one of the seminal works in this space and was learned in a data-driven way from 3D scans of humans.
The SMPL model~\cite{loper2015smpl} and follow-up work~\cite{osman2020star, wang2020blsm, xu2020ghum} streamlined and simplified the body model design making it compatible with traditional graphics pipelines.
A number of extensions of SMPL improved the modeling capabilities, by introducing articulated hands~\cite{romero2017embodied}, facial expressions~\cite{pavlakos2019expressive}, and deformations for the feet~\cite{osman2022supr}.
Although these models increased the detail and realism of the reconstructed surface,
the skeleton maintained the simplistic design of SMPL, representing each body joint with a ball (socket) joint.
This skeleton design was eventually improved by the SKEL model~\cite{keller2023skin}, which adopts most design principles from SMPL, but rigs the surface mesh using a biomechanically accurate skeleton.
For our method, we adopt the SKEL model and estimate its parameters using a single image as input.

\noindent
\textbf{3D Human Pose Estimation.}
Earlier work on 3D human pose estimation was representing the human body with simplistic stick figures~\cite{pavlakos2017coarse,martinez2017simple,sun2018integral}.
Since the introduction of the SMPL model~\cite{loper2015smpl}, there has been a shift towards approaches that reconstruct the full body surface by estimating the parameters of the SMPL model.
Although the initial approaches relied on iterative optimization~\cite{bogo2016keep, pavlakos2019expressive},
currently most methods are based on deep learning and regress the SMPL parameters in a feedforward manner~\cite{kanazawa2018end, pavlakos2018learning}.
HMR~\cite{kanazawa2018end} was a seminal work in this direction that estimates SMPL parameters from a single image with a CNN in an end-to-end manner.
Since then, different designs for the architecture of the network have been proposed~\cite{kocabas2021pare, kocabas2021spec}.
However, most key principles from HMR are still adopted by recent works~\cite{black2023bedlam, goel2023humans}, even when other parametric models are used, like MANO~\cite{romero2017embodied} for hand reconstruction~\cite{pavlakos2024reconstructing} or SMPL-X~\cite{pavlakos2019expressive} for expressive reconstruction~\cite{cai2024smpler, choutas2020monocular}.
One update of recent works~\cite{goel2023humans, cai2024smpler} is the adoption of Visual Transformers~\cite{dosovitskiy2020image, xu2022vitpose} instead of the previous CNN designs~\cite{he2016deep, sun2019deep}.
Following good practices, we also adopt a transformer-based neural network for SKEL regression.

In parallel with the investigation of architecture design for human mesh recovery, other works focused on the data for training.
SPIN~\cite{kolotouros2019learning} proposed an optimization in-the-loop to create pseudo ground truth SMPL parameters for the training images.
EFT~\cite{joo2021exemplar} and CLIFF~\cite{li2022cliff} followed a similar practice with an improved optimization.
In our work, we face the problem that there is no existing image dataset with SKEL ground truth, so we describe how to get an initial dataset with SKEL pseudo ground truth and then iteratively refine these parameters to improve their quality during training.
Besides the data quality, recent work has emphasized the importance of large scale data for training this kind of models~\cite{goel2023humans, cai2024smpler, sarandi2024neural}.
We follow these good practices and we train using the large scale data of HMR2.0~\cite{goel2023humans}.

\noindent
\textbf{Pose Estimation Meets Biomechanics.}
The most common use of human pose estimation methods in biomechanics is in the form of 2D keypoint detectors~\cite{cao2017realtime, xu2022vitpose} that can provide reliable 3D poses after triangulation from multiple views~\cite{pagnon2022pose2sim, uhlrich2023opencap}.
Lin~\etal~\cite{lin20243d} proposed an approach to regress the biomechanical model parameters by using input images from two views, while Bittner~\etal\cite{bittner2022towards} use video input.
In contrast to them, we address the problem in its more challenging, single-image setting.
Jiang~\etal~\cite{jiang2024manikin} use biomechanical constraints for more accurate 3D pose estimation, but their work adopts the SMPL model, making the output incompatible with biomechanical simulations~\cite{delp2007opensim}.
Moreover, there is progress with the datasets for biomechanics.
Werling~\etal~\cite{werling2023addbiomechanics} introduced the AddBiomechanics dataset, a large scale collection of biomechanics data.
This has the potential of acting similarly to the popular AMASS dataset~\cite{mahmood2019amass} enabling training of pose and motion priors.
More recently, Gozlan~\etal~\cite{gozlan2024opencapbench} introduced a benchmark, OpenCapBench, for evaluating human pose estimation methods under physiological constraints.
The benchmark was not available at the time of submission, but it could be useful for evaluating HSMR and future work.
\section{Technical approach}
\label{sec:method}

In this section, we describe our technical approach for reconstructing humans using a biomechanically accurate skeleton model.
First, we provide some preliminaries regarding the SKEL model~\cite{keller2023skin} (Section~\ref{sec:method_preliminaries}), and then we present our HSMR model for Human Skeleton and Mesh Recovery (Section~\ref{sec:method_hsmr}).
We focus on the architecture, the procedure for training data generation, and the iterative refinement of the pseudo ground truth during training.

\subsection{Preliminaries}
\label{sec:method_preliminaries}

\textbf{SKEL Model.}
The SKEL model \cite{keller2023skin} is a parametric body model that combines the popular SMPL model~\cite{loper2015smpl} with a biomechanical skeleton model, BSM.
Specifically, SKEL defines a function $\mathcal{S}(q,\beta)$ that takes as input parameters for pose ($q \in \real^{46}$) and shape ($\beta \in \real^{10}$), and outputs a skin mesh $M \in \real^{3 \times N}$ with $N = 6890$ vertices and a skeleton mesh $S$.
The surface mesh shares the same topology with SMPL,
so we can apply a regressor $W$ to get the locations of the 3D joints $X = WM$.
The shape space of SKEL, and the shape parameters $\beta$ are the same with SMPL.
However, there is a key difference for the pose representation.
Previous models in the SMPL family~\cite{loper2015smpl, romero2017embodied, pavlakos2019expressive} have treated every articulation joint as a ball (socket) joint with three degrees of freedom.
In contrast to that, SKEL carefully designs the kinematic parameters according to the real human biomechanical structure and only models the realistic degrees of freedom.
As a result, the pose parameters $q$ are lower dimensional -- 46 for SKEL, compared to 72 for SMPL.
Each pose parameter corresponds to a single degree of freedom and is represented as an Euler angle.
This allows us to associate each parameter with its explicit joint rotation limits.
For example, the knee has one degree of freedom with limits of 0$^{\circ}$ extension and 135$^{\circ}$ flexion.

\subsection{Human Skeleton and Mesh Recovery}
\label{sec:method_hsmr}

\noindent
\textbf{Architecture.}
For our architecture, we follow best practices from the human mesh recovery literature~\cite{kanazawa2018end, goel2023humans}.
We start with a ViT backbone~\cite{dosovitskiy2020image, xu2022vitpose}, which takes as input an RGB image $I$ of a person.
A transformer head at the end of the network regresses the parameters of the SKEL model.

In terms of the model output, we regress the camera $\pi$, the shape parameters $\beta$, and the pose parameters $q$.
Unlike the SMPL family of models, SKEL represents the pose parameters $q$ with Euler angles.
Although this representation is intuitive,
we find that Euler angles can be challenging as a regression target (Section~\ref{sec:experiments}).
Instead, we adopt the continuous rotation representation~\cite{zhou2019continuity} for the pose parameters.
Initially, the output of the network is in the form of this continuous representation, $q_{\textrm{cont}}$.
We first convert the parameters to the rotation matrix representation, $q_{\textrm{mat}}$, using Gram–Schmidt~\cite{zhou2019continuity}.
The $q_{\textrm{mat}}$ representation is where we apply our parameter loss.
Then, we can convert the parameters to the Euler angle representation, $q_{\textrm{Euler}}$, which is compatible with the input of the SKEL model.
Eventually, the losses on the SKEL parameters are:
\begin{equation}
  \loss_q = ||q_{\textrm{mat}} - q_{\textrm{mat}}^*||_2^2 \textrm{  and  } \loss_{\beta} = ||\beta - \beta^*||_2^2.
\end{equation}
Here, $q_{\textrm{mat}}^*$ and $\beta^*$ are the ground truth pose and shape parameters, respectively.
Besides the parameter losses $\mathcal{L}_q$ and $\mathcal{L}_{\beta}$ (which are applied only when the labels are available),
we also apply losses on the 3D and 2D keypoints:
\begin{equation}
  \label{eqn:keypoints}
  \loss_\texttt{kp3D} = ||X - X^*||_1 \textrm{  and  } \loss_\texttt{kp2D} = ||\pi(X) - x^*||_1.
\end{equation}

\noindent
\textbf{Training Data Generation.}
One key obstacle in training our HSMR model is that there are no image datasets with SKEL annotations.
To address this, we propose to leverage existing image datasets with SMPL (pseudo) ground truth and convert them to SKEL parameters.
This conversion is possible because the two models share the same topology for the surface mesh.
This allows us to optimize the SKEL parameters, such that the SKEL mesh aligns with the target SMPL mesh~\cite{keller2023skin}.
Through this procedure, we can acquire some initial pseudo ground truth SKEL parameters for the datasets typically used for human mesh recovery.

\begin{figure}
    \centering
    \small
    \includegraphics[width=\linewidth]{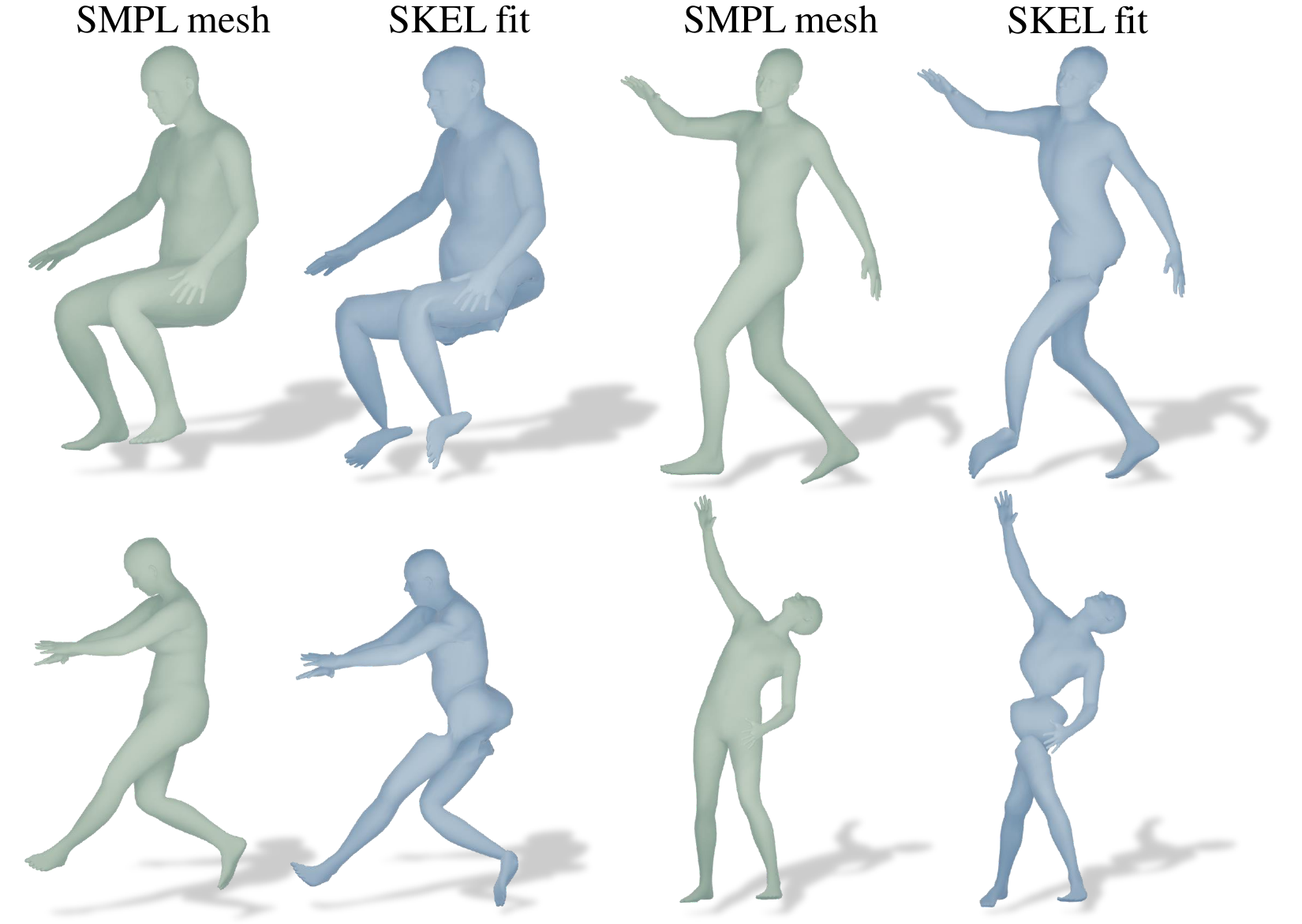}
    \vspace{-0.5em}
    \caption{
    \textbf{Failure cases of SMPL-to-SKEL conversion.}
    While we can technically fit SKEL to an instance of the SMPL model, this conversion can often lead to problematic SKEL results.
    Here, we visualize SMPL meshes (light green), and the SKEL meshes we get when we try to fit the SKEL model to the SMPL mesh (light blue).
    For the fitting, we use the optimization code of~\cite{keller2023skin}.
    }
    \vspace{-1.5em}
    \label{fig:bad_gt_labels}
\end{figure}

\addtolength{\tabcolsep}{-3pt}
\begin{table*}[t]
  \centering
  \footnotesize
  \resizebox{\textwidth}{!}
  {
    \begin{tabular}{cl|cc|cc|cc|cc|cc|cc}
    \toprule
    \multicolumn{2}{c|}{\multirow{2}[4]{*}{Methods}} & \multicolumn{2}{c|}{COCO} & \multicolumn{2}{c|}{LSP-Extended} & \multicolumn{2}{c|}{PoseTrack} & \multicolumn{2}{c|}{3DPW} & \multicolumn{2}{c|}{Human3.6M} & \multicolumn{2}{c}{MOYO} \\
    \cmidrule{3-14}    \multicolumn{2}{c|}{} & @0.05$\uparrow$ & @0.1$\uparrow$ & @0.05$\uparrow$ & @0.1$\uparrow$ & @0.05$\uparrow$ & @0.1$\uparrow$ & MPJPE$\downarrow$ & PA-MPJPE$\downarrow$ & MPJPE$\downarrow$ & PA-MPJPE$\downarrow$ & MPJPE$\downarrow$ & PA-MPJPE$\downarrow$ \\
    \midrule
      & PARE~\cite{kocabas2021pare} & 0.72	& 0.91	& 0.27	& 0.60	& 0.79	& 0.93	& 82.0	& 50.9	& 76.8	& 50.6	& 165.6	& 117.1 \\
      & CLIFF~\cite{li2022cliff} & 0.64	& 0.88	& 0.32	& 0.66	& 0.75	& 0.92	& -- *	& -- *	& 47.1	& 32.7	& 154.6 & 109.3 \\
      & HybrIK~\cite{li2021hybrik} & 0.61	& 0.80	& 0.37	& 0.69	& 0.81	& 0.94 & \textbf{80.0}	& \textbf{48.8}	& 54.4	& 34.5	& 140.1	& 93.2 \\
      & PLIKS~\cite{shetty2023pliks} & 0.62 & 0.90	& 0.26	& 0.66	& 0.74	& 0.94	& -- *	& -- *	& \textbf{47.0}	& 34.5	& 132.6	& 91.8\\
    \midrule
      & HMR2.0~\cite{goel2023humans} & \textbf{0.86}	& \textbf{0.96}	& \textbf{0.53}	& \textbf{0.82}	& \textbf{0.90}	& \textbf{0.98}	& 81.3	& 54.3	& 50.0	& \textbf{32.4}	& 123.3	& 90.4 \\
      & HSMR & \phantom{\bf \tiny +0.01}\hfill          0.85{\raggedleft  \color{soft_red}       \bf \tiny +0.01}	 &
               \phantom{\bf \tiny +0}\hfill     \textbf{0.96}{\raggedleft \color{dark_green} \bf \tiny +0}     &
               \phantom{\bf \tiny +0.02}\hfill          0.51{\raggedleft  \color{soft_red}       \bf \tiny +0.02}	 &
               \phantom{\bf \tiny +0.01}\hfill          0.81{\raggedleft  \color{soft_red}       \bf \tiny +0.01}	 &
               \phantom{\bf \tiny +0}\hfill     \textbf{0.90}{\raggedleft \color{dark_green} \bf \tiny +0}     &
               \phantom{\bf \tiny +0}\hfill     \textbf{0.98}{\raggedleft \color{dark_green} \bf \tiny +0}     &
               \phantom{\bf \tiny +0.2}\hfill           81.5{\raggedleft  \color{soft_red}       \bf \tiny +0.2}   &
               \phantom{\bf \tiny +0.5}                 54.8{\raggedleft  \color{soft_red}       \bf \tiny +0.5}\; &
               \phantom{\bf \tiny +0.4}\hfill           50.4{\raggedleft  \color{soft_red}       \bf \tiny +0.4}   &
               \phantom{\bf \tiny +0.5}                 32.9{\raggedleft  \color{soft_red}       \bf \tiny +0.5}\; &
               \phantom{\bf \tiny -18.8}\hfill \textbf{104.5}{\raggedleft \color{dark_green} \bf \tiny -18.8}  &
               \phantom{\bf \tiny -10.8}        \textbf{79.6}{\raggedleft \color{dark_green} \bf \tiny -10.8}\;\\
    \bottomrule
    \end{tabular}
  }
  \vspace{-0.5em}
  \caption{\textbf{Comparison with state-of-the-art approaches that regress SMPL parameters.}
  The primary baseline for HSMR is the HMR2.0 network~\cite{goel2023humans}, since it is the closest to our design, in terms of architecture and training data
  We report PCK @0.05 \& @0.1 for the 2D datasets (COCO, LSP-Extended, PoseTrack) and MPJPE \& PA-MPJPE for the 3D datasets (3DPW, Human3.6M, MOYO).
  Even though we adopt the SKEL model which is less flexible and we start without any initial ground truth for training, we are able to match the performance of HMR2.0 on most datasets - with up to 0.5mm difference.
  More importantly, we outperform HMR2.0 by a big gap of more than 10mm on the challenging MOYO dataset that includes extreme poses and viewpoints.
  In the table, we explicitly report the differences in evaluation metrics between our HSMR network and HMR2.0.
  *: trains on 3DPW.
  }
  \vspace{-1.2em}
  \label{tab:smpl_recon}%
\end{table*}%
\addtolength{\tabcolsep}{3pt}

\noindent
\textbf{Training with Pseudo-Label Refinement.}
Although the SMPL-to-SKEL conversion gives us a reasonable starting point, it is an imperfect procedure with frequent failure cases (Figure~\ref{fig:bad_gt_labels}).
This type of local minima are common in similar iterative optimization problems~\cite{bogo2016keep, pavlakos2019expressive}.
If we aim to improve the accuracy of HSMR, we need to improve the quality of the pseudo ground truth we use for training.

To achieve this, we propose an iterative procedure that gradually updates the quality of the pseudo ground truth SKEL parameters for each example.
This is inspired by previous work on pseudo ground truth refinement~\cite{joo2021exemplar, kolotouros2019learning}.
More specifically, for each image $I$ of a person, given a network estimate $q^{\textrm{reg}}, \beta^{\textrm{reg}}$,
we refine the parameters iteratively, such that they align with the 2D keypoints $x^*$ of the person on the image~\cite{bogo2016keep, pavlakos2019expressive}.
The optimized estimates of the pose and shape parameters, $q^*, \beta^*$ are used as more accurate pseudo ground truth for supervising the network.

For this iterative optimization, we propose an equivalent of SMPLify~\cite{bogo2016keep} for SKEL, which we call SKELify.
The optimization is mainly guided by the 2D keypoints $x^*$.
Specifically, we introduce a reprojection objective, $E_\texttt{kp2D}$, aiming to align the projection of the 3D joints with the 2D keypoints.
This objective is similar to the second part of Equation~\ref{eqn:keypoints}, with the addition of a robustifier~\cite{geman1987statistical} as in~\cite{bogo2016keep}.
To regularize the shape and pose parameters we add shape and pose priors.
The shape prior is inherited from SMPL, \ie, $E_\texttt{shape}(\beta) = \|\beta\|^2$.
For the pose parameters, however, we do not have an existing pose prior for SKEL.
Instead, we leverage the known limits of natural rotation for each joint. %, which are known from biomechanics.
For example, let us assume that for a pose parameter $q_i$, the lower limit is $l_i$ and the upper limit is $u_i$, \ie, $q_i \in [l_i, u_i]$.
In this case, we can add a term:
\begin{equation}
E_\texttt{pose}(q) = \sum_i \exp(l_i - q_i) + \exp(q_i - u_i),
\end{equation}
which strongly penalizes rotations that exceed the known joint limits.
If for a specific parameter there is no explicit limit, we can omit it from the calculation of the objective.

In the end, we sum the three objectives, $E_\texttt{kp2D}(q, \beta)$, $E_\texttt{shape}(\beta)$ and $E_\texttt{pose}(q)$ and solve for the optimal SKEL parameters, $q^*, \beta^*$.
These parameters are used as pseudo ground truth to train the network.
Unlike~\cite{kolotouros2019learning}, this refinement is not happening in every training iteration, but we execute it periodically in batch mode for efficiency reasons. We refer to the SuppMat for more implementation details.
\section{Experiments}
\label{sec:experiments}

\subsection{Datasets and Metrics}
We train HSMR using the training data from HMR2.0~\cite{goel2023humans},
which include images from Human3.6M~\cite{ionescu2013human3}, MPI-INF-3DHP~\cite{mehta2017monocular}, COCO~\cite{lin2014microsoft}, MPII~\cite{andriluka20142d},
AI Challenger~\cite{wu2017ai}, AVA~\cite{gu2018ava} and InstaVariety~\cite{kanazawa2019learning}.
We preprocess the data to convert the SMPL (pseudo) ground truth of HMR2.0 to SKEL parameters, as we describe in Section~\ref{sec:method_hsmr}.

We evaluate our approach on multiple datasets for human pose estimation.
Some of them provide 3D annotations, \ie, Human3.6M~\cite{ionescu2013human3}, 3DPW~\cite{von2018recovering} and MOYO~\cite{tripathi20233d},
while others only include 2D annotations, \ie, COCO~\cite{lin2014microsoft}, PoseTrack~\cite{andriluka2018posetrack} and
LSP Extended~\cite{johnson2011learning}.
Accordingly, we report Percentage of Correct Keypoints (PCK)~\cite{yang2012articulated} at different thresholds as metrics for 2D pose accuracy,
and Mean Per Joint Position Error (MPJPE)~\cite{ionescu2013human3}, Mean Per Vertex Position Error (MPVPE)~\cite{pavlakos2019expressive}
plus their Procrustes Alignment version PA-MPJPE~\cite{kanazawa2018end, zhou2018monocap} and PA-MPVPE as 3D pose accuracy metrics.
Moreover, we evaluate the results of different methods in terms of violation of the joint limits.
Specifically, we focus on knees and elbows and report the frequency of violation for different angle thresholds.
Please see the the SuppMat for more details.

\addtolength{\tabcolsep}{-2pt}
\begin{table}[!t]
  \centering
  \scriptsize
  {
  \begin{tabular}{r|c|c|c|c|c|c}
  \toprule
  & PARE & CLIFF & HybrIK & PLIKS & HMR2.0 & HSMR \\
  \midrule
          MPVPE$\downarrow$ & 174.5	& 155.7 & 143.6	& 136.7	& 142.2	& {\bf 120.1} \\
          PA-MPVPE$\downarrow$ & 121.9 & 110.6 & 94.4 & 94.8	& 103.4	& {\bf 90.7} \\
  \bottomrule
  \end{tabular}
  }
  \vspace{-0.5em}
  \caption{
      {\bf Evaluation of the surface reconstruction accuracy.}
      We report MPVPE and PA-MPVPE on the MOYO dataset.
    }
  \label{tab:vertex_errors}%
  \vspace{-2em}
\end{table}%
\addtolength{\tabcolsep}{2pt}

\addtolength{\tabcolsep}{-2pt}
\begin{table*}[t]
  \centering
  \footnotesize
  \scriptsize
{
    \begin{tabular}{cl|cc|cc|cc|cc|cc|cc}
    \toprule
    \multicolumn{2}{c|}{\multirow{2}[4]{*}{Methods}} & \multicolumn{2}{c|}{COCO} & \multicolumn{2}{c|}{LSP-Extended} & \multicolumn{2}{c|}{PoseTrack} & \multicolumn{2}{c|}{3DPW} & \multicolumn{2}{c|}{Human3.6M} & \multicolumn{2}{c}{MOYO} \\
    \cmidrule{3-14}    \multicolumn{2}{c|}{} & @0.05$\uparrow$ & @0.1$\uparrow$ & @0.05$\uparrow$ & @0.1$\uparrow$ & @0.05$\uparrow$ & @0.1$\uparrow$ & MPJPE$\downarrow$ & PA-MPJPE$\downarrow$ & MPJPE$\downarrow$ & PA-MPJPE$\downarrow$ & MPJPE$\downarrow$ & PA-MPJPE$\downarrow$ \\
    \midrule
          & HMR2.0~\cite{goel2023humans} & \textbf{0.86}	& \textbf{0.96}	& \textbf{0.53}	& \textbf{0.82}	& \textbf{0.90}	& \textbf{0.98}	& 81.3	& \textbf{54.3}	& \textbf{50.0}	& \textbf{32.4}	& 123.3	& 90.4 \\
          & HMR2.0 + SKEL fit & 0.78	& 0.95	& 0.49	& 0.79	& \textbf{0.90}	& \textbf{0.98}	& \textbf{81.0}	& 54.4	& 53.6	& 34.1	& 130.5	& 93.7 \\
          & HSMR & 0.85	& \textbf{0.96}	& 0.51	& 0.81	& \textbf{0.90}	& \textbf{0.98}	& 81.5	& 54.8	& 50.4	& 32.9	& \textbf{104.5}	& \textbf{79.6} \\
    \bottomrule
    \end{tabular}
    }
    \vspace{-0.5em}
    \caption{\textbf{Comparison with baseline for SKEL recovery.}
    We start from the SMPL prediction of HMR2.0~\cite{goel2023humans} and we fit the SKEL model to it with terative optimization~\cite{keller2023skin}.
    This baseline corresponds to the ``HMR2.0 + SKEL fit'' row.
    We observe that this two-stage baseline for SKEL recovery performs worse than HSMR, while it is also significantly slower (3 minutes for a single frame).
    }
    \vspace{-1.5em}
  \label{tab:skel_recon}%
\end{table*}%
\addtolength{\tabcolsep}{2pt}

\subsection{Comparison with methods for SMPL recovery}
\label{sec:exp_smpl_recon}

We build HSMR using best practices from the methods that regress SMPL parameters.
More specifically, HMR2.0~\cite{goel2023humans} is closer to our design, so this is the primary baseline we compare against.
In Table~\ref{tab:smpl_recon}, we compare the performance of HSMR and HMR2.0 on various datasets.
For context, we also include other state-of-the-art methods for SMPL reconstruction~\cite{kocabas2021pare, li2022cliff, li2021hybrik, shetty2023pliks}.
In addition, in Table~\ref{tab:vertex_errors}, we also present results on MOYO for per-vertex errors.

We observe that for most datasets, HSMR achieves results that are almost identical to HMR2.0, with the metrics in 3DPW and Human3.6M having a difference of {\it up to 0.5mm}.
This is important, because even though we operate with a less flexible model (SKEL) and we started our investigation without any initial ground truth for training, we were able to actually match the performance of HMR2.0.
Moreover, we observe that simultaneously we achieve a huge improvement of {\it more than 10mm} on the MOYO dataset~\cite{tripathi20233d}.
The observations are similar for the surface-based evaluation (Table~\ref{tab:vertex_errors}).
This is significant, because MOYO includes challenging extreme poses (yoga poses) and viewpoints.
We believe that this could be attributed to the stronger pose regularization that the biomechanical skeleton can impose, since it only allows the realistic degrees of freedom.
In fact, in Section~\ref{sec:exp_violation}, we verify that the various networks regressing SMPL parameters are indeed suffering from frequent violations of the joint limits.

\subsection{Baseline for SKEL recovery}

Besides comparing with methods for SMPL-based reconstruction, we also consider an optimization-based baseline for SKEL reconstruction.
This was introduced by~\cite{keller2023skin}
and it is the same with the approach we use for our pseudo ground truth generation (Section~\ref{sec:method_hsmr}).
For the comparison, we run HMR2.0 to get SMPL parameters and we fit SKEL to the SMPL mesh with the optimization approach.
The full results are presented in Table~\ref{tab:skel_recon}.
Although in some cases the SKEL fit is comparable with the HMR2.0 output (\eg, PoseTrack and 3DPW), in most cases there is a clear degradation in the quality (\ie, COCO, LSP-Extended, Human3.6M and MOYO).
Additionally, the fitting procedure is computationally expensive, requiring 3 minutes per frame.
This means that our end-to-end HSMR approach is not only more accurate, but also much faster than the SKEL fitting.

\subsection{Biomechanically-sound reconstruction}
\label{sec:exp_violation}

\begin{figure}
    \centering
    \small
    \includegraphics[width=\linewidth]{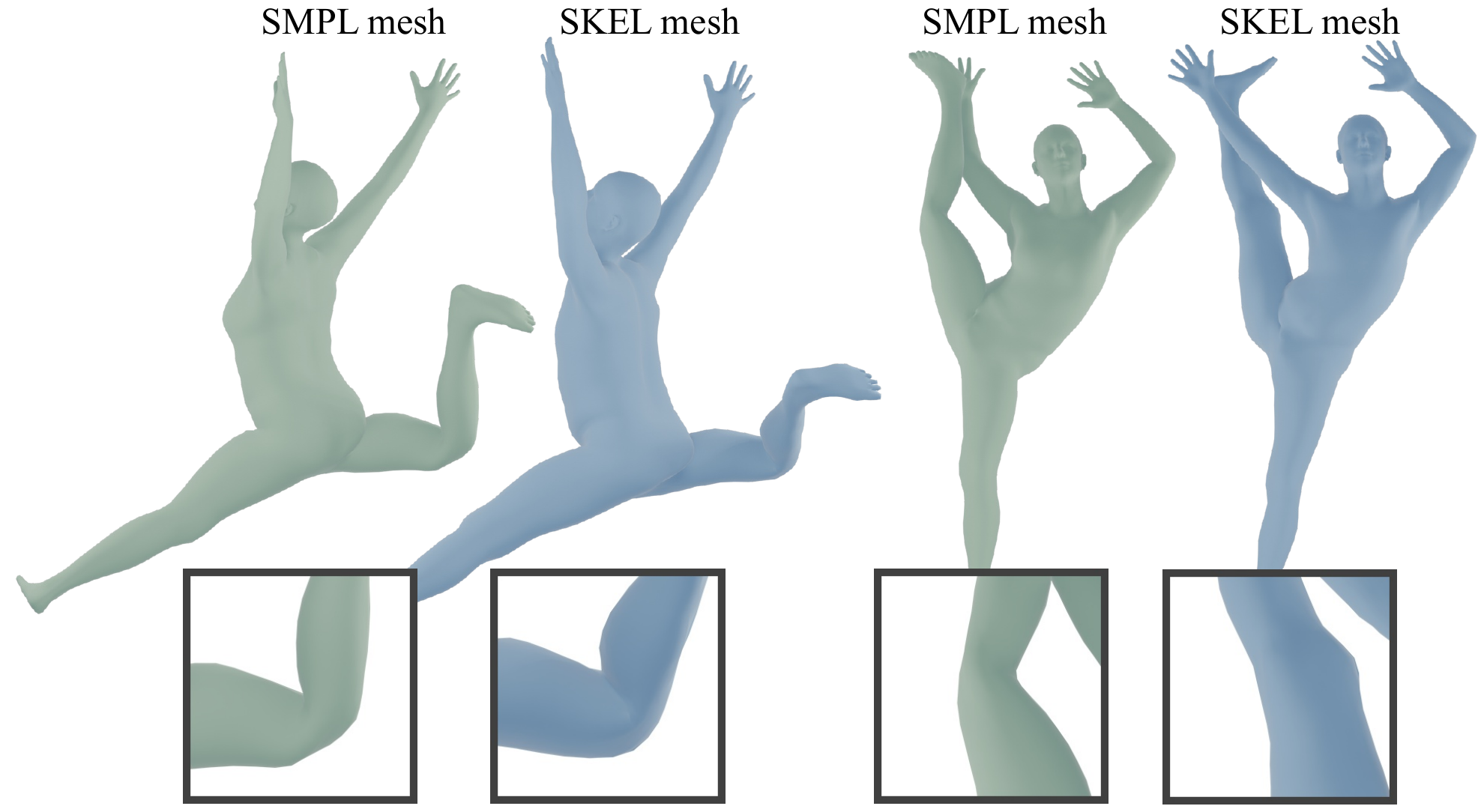}
    \vspace{-1.5em}
    \caption{
    \textbf{Examples of unnatural joint rotation for SMPL.}
    SMPL represents the knee with a ball (socket) joint.
    This allows mesh recovery methods like HMR2.0~\cite{goel2023humans} to generate invalid rotations.
    We visualize examples from HMR2.0 (light green) where the knee is bend in unnatural ways.
    In comparison, the HSMR output (light blue) respects the biomechanical constraints.
    }
    \label{fig:invalid_rotation}
    \vspace{-1.5em}
\end{figure}

Besides evaluating the 2D/3D pose accuracy of the different mesh recovery approaches, we also investigate the biomechanical validity of their outputs.
As discussed in Section~\ref{sec:method_preliminaries}, SKEL only considers the realistic degrees of freedom for each joint, whereas SMPL models each joint with a ball (socket) joint, which endows three degrees of freedom for each joint.
In this subsection, we investigate whether methods that regress SMPL parameters actually predict unnatural joint rotations.
We focus our attention specifically on the elbow and the knee joints.
We consider various thresholds (\ie, 10$^\circ$, 20$^\circ$, 30$^\circ$) and report the frequency that each method exceeds this threshold (\ie, rotation violation).
The complete results for MOYO are presented in Table~\ref{tab:violation}.
As we can see, the violations are more frequent than we might have expected and they happen for all the methods that regress SMPL parameters.
These results are an indication that these methods might return poses with low 3D joint position errors that rotate the body parts in unnatural ways.
We visualize some interesting failure cases in Figure~\ref{fig:invalid_rotation}.
We believe this observation points to a clear direction for future improvement of the approaches for human mesh recovery.

\addtolength{\tabcolsep}{-5pt}
\begin{table*}[t]
  \centering
  \footnotesize
  \resizebox{0.9\textwidth}{!}{
    \begin{tabular}{cl|cccc|cccc|cccc}
    \toprule
    \multicolumn{2}{c|}{\multirow{2}[4]{*}{Methods}} & \multicolumn{4}{c|}{violation $> 10^\circ\downarrow$} & \multicolumn{4}{c|}{violation $> 20^\circ\downarrow$} & \multicolumn{4}{c}{violation $> 30^\circ \downarrow$}  \\
    \cmidrule{3-14}    \multicolumn{2}{c|}{} & \textrm{left elbow} & \textrm{right elbow} & \textrm{left knee} & \textrm{right knee} & \textrm{left elbow} & \textrm{right elbow} & \textrm{left knee} & \textrm{right knee} & \textrm{left elbow} & \textrm{right elbow} & \textrm{left knee} & \textrm{right knee} \\
    \midrule
          & PARE~\cite{kocabas2021pare} & 36.4\%	& 42.4\%	& 20.0\%	& 23.2\%	& 14.6\%	& 15.4\%	& 3.2\%	& 3.8\%	& 5.5\%	& 4.8\%	& 0.3\%	& 0.4\% \\
          & CLIFF~\cite{li2022cliff} & 34.2\%	& 33.0\%	& 28.3\%	& 31.0\%	& 13.0\%	& 12.4\%	& 4.8\%	& 4.5\%	& 5.2\%	& 5.2\%	& 0.5\%	& 0.3\% \\
          & HybrIK &	58.7\%	 & 	60.9\%	 & 	52.9\%	 & 	48.6\%	 & 	29.4\%	 & 	34.6\%	 & 	30.7\%	 & 	27.0\%	 & 	16.4\%	 & 	21.0\%	 & 	20.0\%	 & 	17.5\%	\\
          & PLIKS & 41.6\%	 & 44.7\%	 & 47.4\%	 & 43.8\%	 & 17.9\%	 & 22.7\%	 & 18.2\%	 & 17.6\%	 & 8.3\%	 & 11.4\%	 & 8.5\%	 & 8.5\%	\\
          & HMR2.0~\cite{goel2023humans} & 47.6\%	& 44.3\%	& 45.7\%	& 56.4\%	& 19.8\%	& 19.6\%	& 6.4\%	& 11.6\%	& 8.5\%	& 8.8\%	& 1.0\%	& 1.6\% \\
          & HSMR & {\bf 0.0\%}	& {\bf 0.0\%}	& {\bf 3.9\%}	& {\bf 4.5\%}	& {\bf 0.0\%}	& {\bf 0.0\%}	& {\bf 0.2\%}	& {\bf 0.5\%}	& {\bf 0.0\%}	& {\bf 0.0\%}	& {\bf 0.0\%}	& {\bf 0.0\%}	\\
    \bottomrule
    \end{tabular}
    }
    \vspace{-0.5em}
    \caption{\textbf{Frequency of unnatural rotations for mesh recovery approaches.}
    We investigate how often each approach returns 3D bodies with unnatural joint rotations.
    We experiment on MOYO~\cite{tripathi20233d} and
    report the frequency that the unnatural rotation exceeds different thresholds ( 10$^\circ$, 20$^\circ$ or 30$^\circ$) for the elbow and the knee joints.
    Methods that regress SMPL parameters violate the joint limits frequently.
    Instead, our HSMR method avoids severe violations because it relies on SKEL which models only the realistic degrees of freedom.
    }
    \vspace{-0.5em}
  \label{tab:violation}%
\end{table*}%
\addtolength{\tabcolsep}{5pt}

\begin{figure*}
    \centering
    \includegraphics[width=0.98\textwidth]{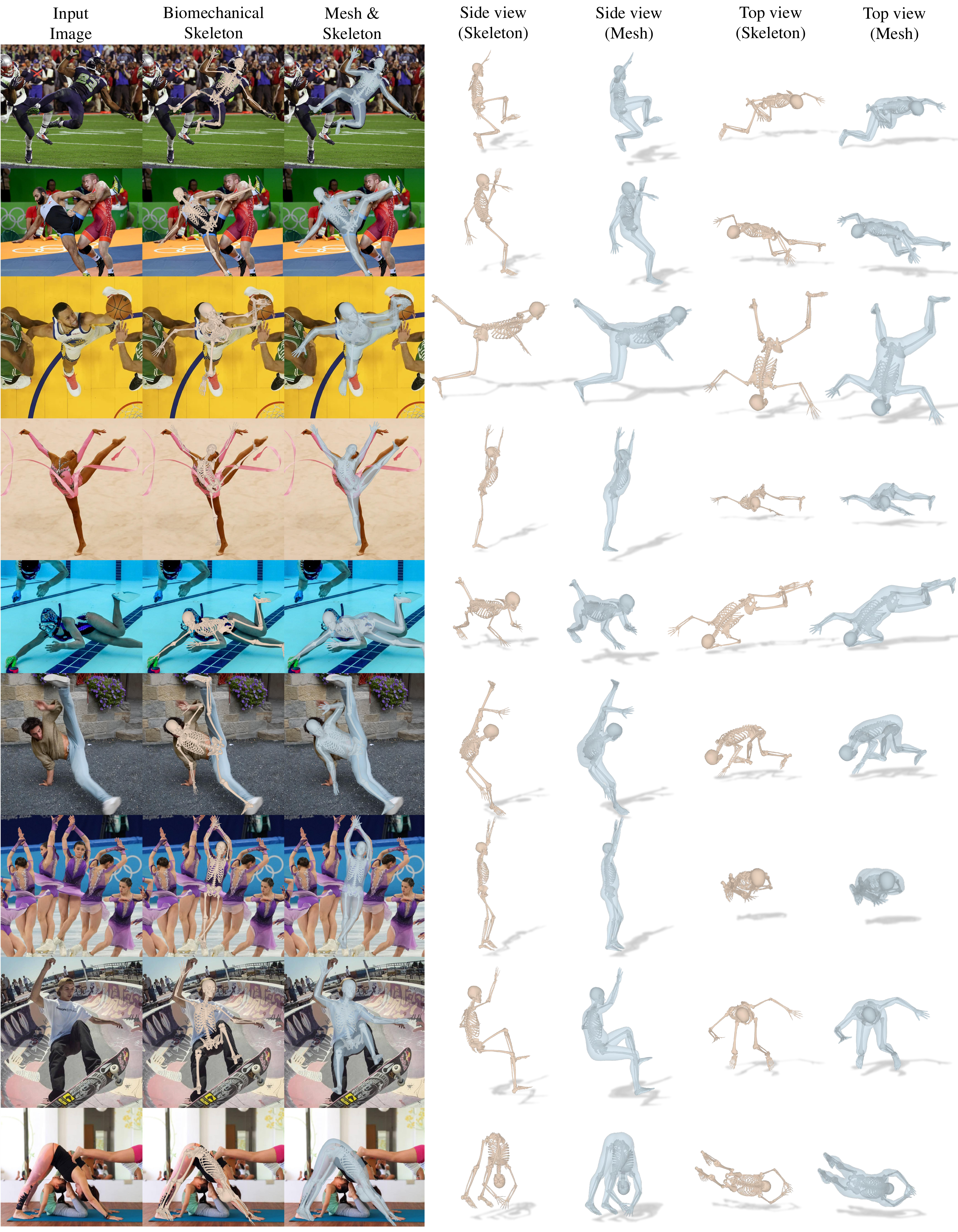}
    \vspace{-0.5em}
    \caption{{\bf Qualitative evaluation of HSMR.}
    For each input example we show:
    a) the input image,
    b) the overlay of SKEL in the input view,
    c) a side view,
    d) the top view.
    We visualize both the skeleton and the transparent mesh of the estimated SKEL.
    }
    \label{fig:qualitative}
    \vspace{-0.5em}
\end{figure*}

\addtolength{\tabcolsep}{-5pt}
\begin{table*}[t]
\centering
\footnotesize
\resizebox{\textwidth}{!}{
  \begin{tabular}{cl|cc|cc|cc|cc|cc|cc}
    \toprule
    \multicolumn{2}{c|}{\multirow{2}[4]{*}{Models}} & \multicolumn{2}{c|}{COCO} & \multicolumn{2}{c|}{LSP-Extended} & \multicolumn{2}{c|}{PoseTrack} & \multicolumn{2}{c|}{3DPW} & \multicolumn{2}{c|}{Human3.6M} & \multicolumn{2}{c}{MOYO} \\
    \cmidrule{3-14}    \multicolumn{2}{c|}{} & @0.05$\uparrow$ & @0.1$\uparrow$ & @0.05$\uparrow$ & @0.1$\uparrow$ & @0.05$\uparrow$ & @0.1$\uparrow$ & MPJPE$\downarrow$ & PA-MPJPE$\downarrow$ & MPJPE$\downarrow$ & PA-MPJPE$\downarrow$ & MPJPE$\downarrow$ & PA-MPJPE$\downarrow$ \\
    \midrule
        & HSMR (ViT-B) & {\bf 0.79}	& {\bf 0.94}	& {\bf 0.38}	& {\bf 0.70}	& {\bf 0.86}	& {\bf 0.96} & {\bf 76.7}	& {\bf 50.0}	& {\bf 49.8}	& {\bf 37.1}	& {\bf 124.0}	& {\bf 92.6} \\
        & HSMR (ViT-B) w/ Euler angles & 0.75 & 0.93	& 0.31	& 0.64	& 0.82	& 0.95	& 81.6	& 52.1	& 55.6	& 41.3	& 137.1	& 104.3 \\
        & HSMR (ViT-B) w/o pseudo GT refinement & 0.75 & 0.93	& 0.37	& {\bf 0.70}	& 0.84	& {\bf 0.96}	& 81.1	& 51.1	& 52.0	& 38.1	& 126.5	& 96.2 \\
    \bottomrule
  \end{tabular}
}
\vspace{-0.5em}
\caption{
  \textbf{Ablation study on design choices.}
  We benchmark our proposed model and ablate two design choices.
  First, we change the regression target from the continuous representation~\cite{zhou2019continuity} to the native Euler angles of SKEL. This has a negative effect across the board.
  Then, we experiment without the pseudo ground truth refinement process.
  This also has a negative impact particularly on the 3D metrics.
}
\vspace{-1.0em}
\label{tab:ablation}%
\end{table*}%
\addtolength{\tabcolsep}{5pt}

\subsection{Ablation study}
Finally, we evaluate some key design decisions of our pipeline.
More specifically, we investigate the choice of regression target for the pose parameters.
We compare using the continuous rotation representation~\cite{zhou2019continuity} as an alternative to the
Euler angles (which is the native representation for SKEL).
Moreover, we assess the importance of iterative refinement of the SKEL pseudo ground truth that we employ during training.
For this evaluation, we perform a smaller scale ablation using a ViT-B backbone~\cite{xu2022vitpose} for our network.

We present the detailed results of this ablation in Table~\ref{tab:ablation}.
As we see, regressing the Euler angles directly produces a clear drop in performance,
justifying the use of the continuous rotation representation for SKEL parameter regression.

Moreover, if we train without the iterative refinement of the labels, the performance decreases for most datasets, particularly for the 3D metrics (for the 2D metrics, the difference is small, because the refinement does not affect the quality of the 2D pseudo ground truth).
These results confirm the importance of both design choices.

\subsection{Qualitative evaluation}

\begin{figure}
    \centering
    \includegraphics[width=0.98\columnwidth]{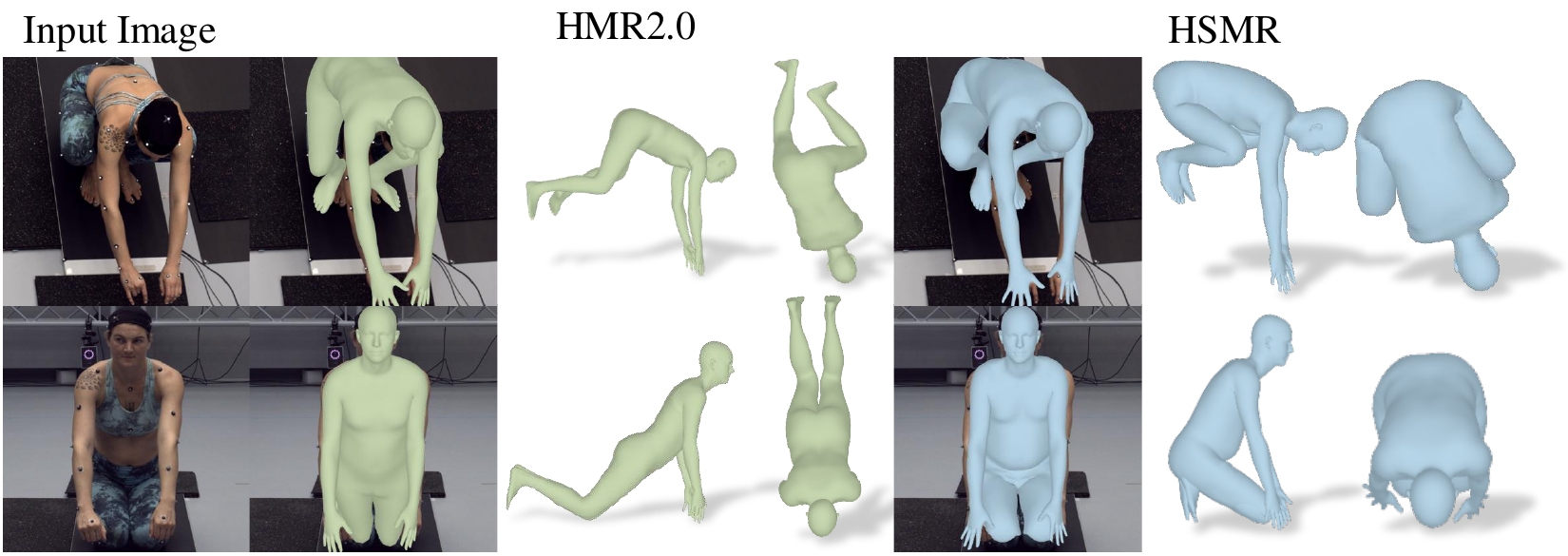}
    \vspace{-0.5em}
    \caption{{\bf Qualitative comparison with HMR2.0 on MOYO.}
    For each example we show the input image and results for HMR2.0 and HSMR.
    Although the interpretation in the input view is reasonable for both methods, HSMR achieves more accurate 3D reconstruction on the challenging poses and viewpoints of MOYO.
    }
    \label{fig:qualitative_hmr2}
    \vspace{-1.2em}
\end{figure}

\begin{figure}[!t]
    \centering
    \small
    \includegraphics[width=\linewidth]{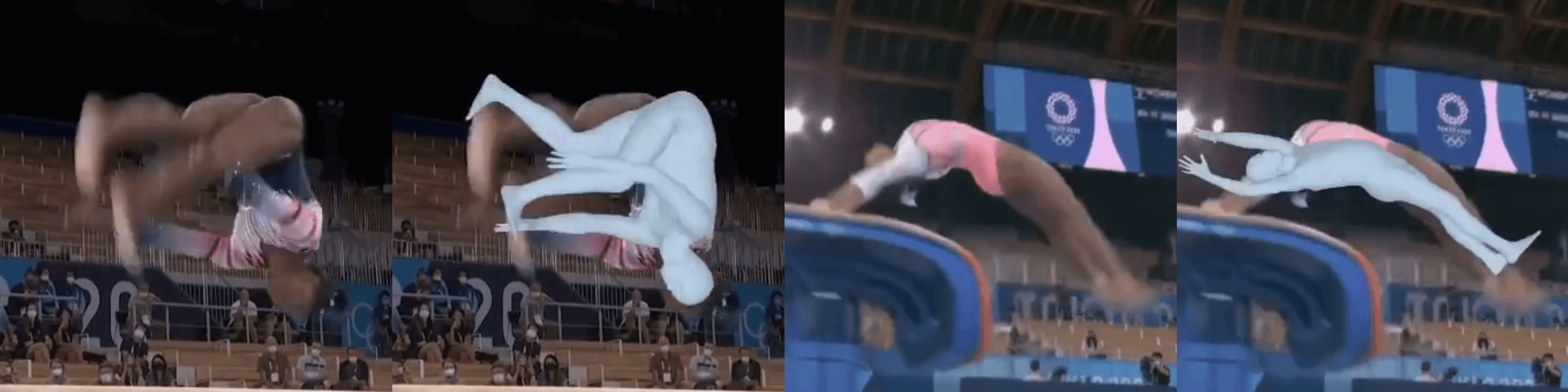}
    \includegraphics[width=\linewidth]{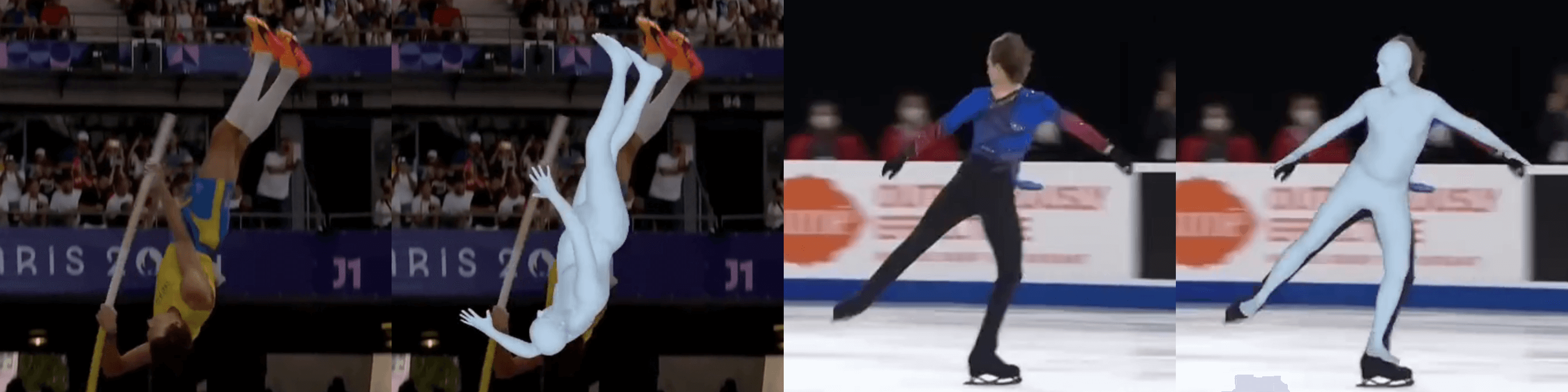}
    \vspace{-1.5em}
    \caption{
        {\bf Failure cases of our method.}
        HSMR often fails in cases with motion blur extreme poses and rare viewpoints.
    }
    \vspace{-2.0em}
    \label{fig:supmat_failure}
\end{figure}

In Figure~\ref{fig:qualitative}, we provide more qualitative results of our approach. We show reprojections on the image, as well as side and top views.
We visualize both the (transparent) surface mesh and the skeleton output.
HSMR performs well for a variety of poses, and viewpoints.
Also, in Figure~\ref{fig:qualitative_hmr2} we show a comparison with HMR2.0 on images from the MOYO dataset.
The qualitative improvements achieved by HSMR align with the MOYO quantitative results of Table~\ref{tab:smpl_recon}.
Finally, in Figure~\ref{fig:supmat_failure}, we present some failure cases of HSMR.
\section{Summary}
\label{sec:summary}

In this paper, we presented an approach for reconstructing humans in 3D using a biomechanically accurate model, SKEL.
We design a network that takes a single image as input and estimates the parameter of the SKEL model.
To achieve that, we curate existing datasets with pseudo ground truth SKEL parameters and use them to train our model.
In terms of 3D body pose estimation,
our approach matches the performance of the state-of-the-art human mesh recovery methods while also outperforming them on cases with challenging poses and uncommon viewpoints.
Moreover, we demonstrate how previous approaches for SMPL regression are failing to respect the biomechanical constraints, leading to serious violations of the joint angle limits.
We hope that our work will help close the gap between vision-based methods for human pose estimation and the high precision required for biomechanical analysis.

\noindent
\textbf{Limitations and future work.}
One of the limitations of HSMR is the exclusive use of pseudo ground truth for training.
Although our iterative refinement improves the pseudo ground truth quality,
the network could benefit from more precise 3D labels.
Moreover, we observe some inevitable jitter in our temporal reconstructions.
We believe that follow-up work could address the recovery of smooth SKEL motions.
Finally, future work could consider incorporating our estimates in a biomechanical simulation environment~\cite{delp2007opensim} to encourage physically-plausible motion~\cite{ugrinovic2024multiphys}.

\footnotesize
\noindent
\textbf{Acknowledgements:}
E.V. was supported by CMMI-2310666.
X.Z. was supported by Zhejiang Provincial Natural Science Foundation of China (No. LR25F020003) and Information Technology Center and State Key Lab of CAD\&CG, Zhejiang University.
Q.H. was supported by NSF IIS-2047677, NSF IIS-2413161, and Gifts from Adobe and Google.
G.P. was supported by Gifts from Google and Adobe.

{
    \small
    \bibliographystyle{ieeenat_fullname}
    \bibliography{main}
}

\end{document}